%% file: main.tex
\documentclass[letterpaper]{article} 
\usepackage[preprint]{aaai2027}  
\usepackage[hyphens]{url}  
\usepackage{graphicx} 
\usepackage{natbib}  
\usepackage{caption} 
\usepackage{booktabs}
\usepackage{multirow}
\usepackage{tabularx}
\usepackage{amssymb}
\usepackage{amsmath}
\usepackage[table]{xcolor}
\usepackage{tcolorbox}
\definecolor{graycol}{gray}{0.90}

\newcommand{\cmark}{\checkmark}
\newcommand{\xmark}{\(\times\)}

\title{DiffCap-Bench: A Comprehensive, Challenging, Robust Benchmark for Image Difference Captioning}

\author{
Yuancheng Wei$^{\heartsuit}$\equalcontrib\quad
Haojie Zhang$^{\heartsuit}$\equalcontrib\quad
Linli Yao$^{\diamond}$\quad
Lei Li$^{\spadesuit}$\\
Jiali Chen$^{\heartsuit}$\quad
Tao Huang$^{\clubsuit}$\quad
Yiting Lu$^{\ddagger}$\\
Duojun Huang$^{\ddagger}$\quad
Xin Li$^{\ddagger}$\corresponding\quad
Zhao Zhong$^{\ddagger}$
}
\affiliations{
$^{\heartsuit}$South China University of Technology\\
$^{\diamond}$Peking University\\
$^{\spadesuit}$The University of Hong Kong\\
$^{\clubsuit}$Tianjin University\\
$^{\ddagger}$Tencent Hunyuan
}

\makeatletter
\let\aaai@orig@maketitle\@maketitle
\def\@maketitle{%
  \aaai@orig@maketitle
  \vspace{0.4em}
  \centering
  \scalebox{1}[0.95]{\includegraphics[width=0.82\textwidth]{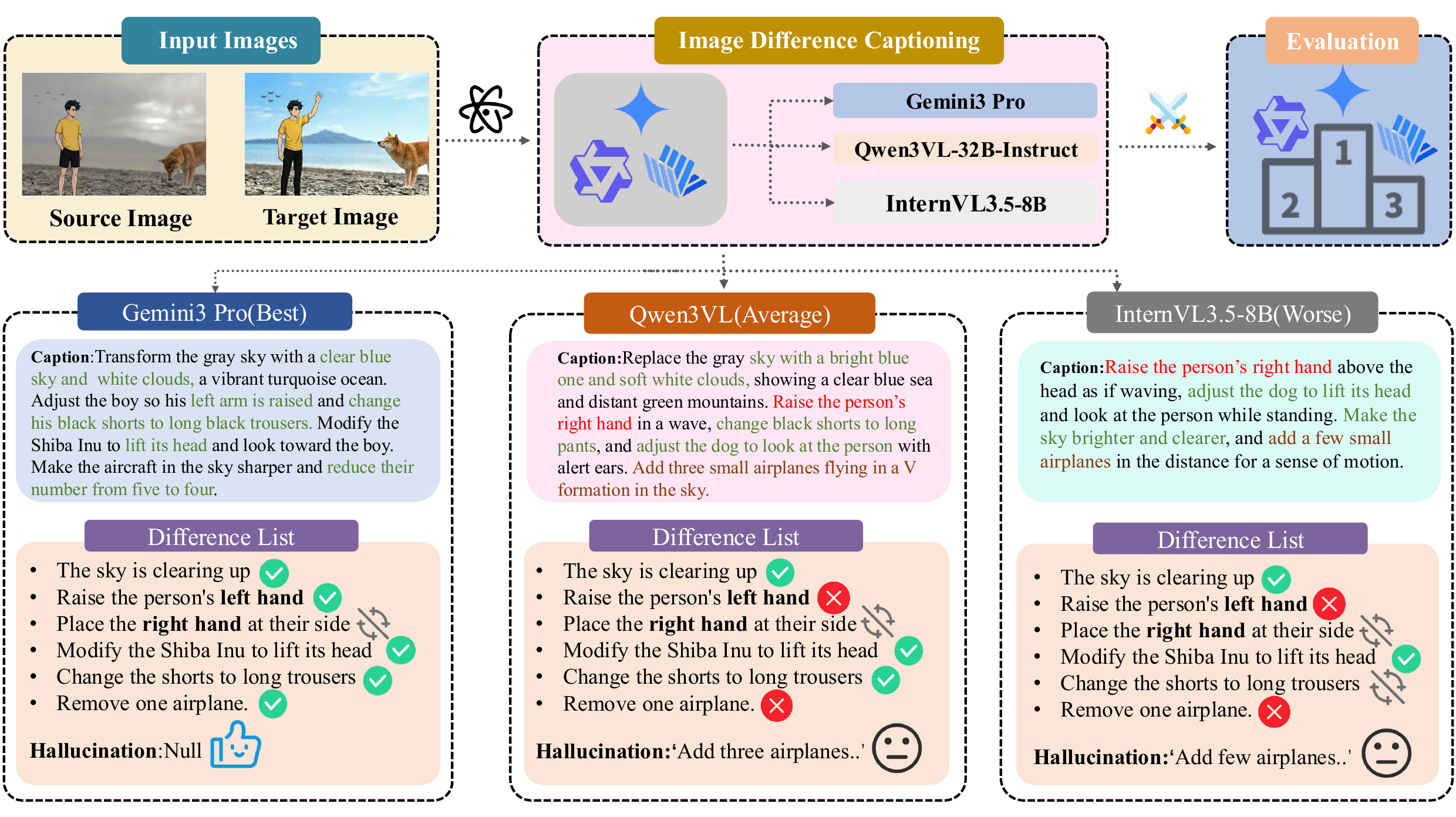}}
  \captionof{figure}{An example from DiffCap-Bench. Captions are checked against a human-validated Difference List:
  (1)~\textbf{\textcolor{green}{\checkmark}} correctly described;
  (2)~\textbf{\textcolor{red}{$\times$}} incorrectly described;
  (3)~\textbf{\textcolor{gray}{Missed}} omitted;
  \textbf{Hallucination} denotes fabricated changes.
  \textbf{Gemini 3 Pro} captures differences with few hallucinations, while open-source models show more errors.}
  \label{fig:teaser}
  \vspace{0.5em}
}
\makeatother

\begin{document}

\maketitle

\begingroup
\renewcommand{\thefootnote}{}
\footnotetext{
\pdfstartlink attr{/Border [0 0 0]} user{
/Subtype /Link /A << /S /URI /URI (mailto:wyc528813339@gmail.com) >>}
\texttt{wyc528813339@gmail.com}\pdfendlink\\
\hspace*{1.8em}%
\pdfstartlink attr{/Border [0 0 0]} user{
/Subtype /Link /A << /S /URI /URI (https://github.com/wyclike/DiffCap-Bench) >>}
\url{https://github.com/wyclike/DiffCap-Bench}\pdfendlink}
\endgroup

\begin{abstract}
\input{sections/abstract}
\end{abstract}

\section{Introduction}
\label{sec:intro}

\input{sections/introduction}

\section{Related Work}
\label{sec:related}

\begin{figure*}[t]
\centering
\includegraphics[width=0.82\textwidth]{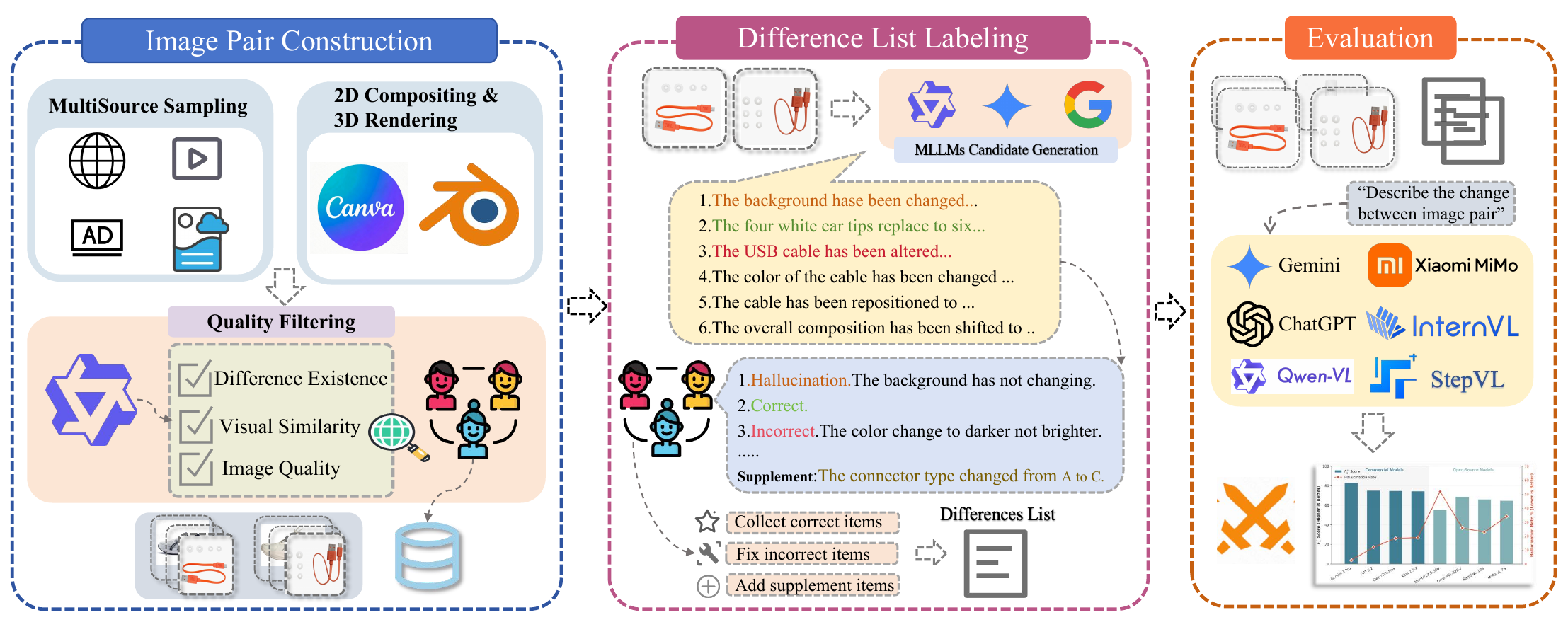}
\caption{Overview of the DiffCap-Bench construction pipeline, consisting of three stages: \emph{Image Pair Construction} collects and filters image pairs from diverse sources; \emph{Difference List Labeling} builds a human-validated \emph{Difference List} of atomic difference items per pair; and \emph{Evaluation} systematically benchmarks current MLLMs to reveal their IDC capabilities and limitations.}
\label{fig:pipeline}
\end{figure*}

\input{sections/relatedworks}

\section{DiffCap-Bench}
\label{sec:DiffCap-Bench}

\begin{table}[t]
  \centering
  \scriptsize
  \setlength{\tabcolsep}{3pt}
  \renewcommand{\arraystretch}{0.88}
  \begin{tabularx}{\columnwidth}{@{} l r X @{}}
    \toprule
    \textbf{Entry} & \textbf{Count} & \textbf{Definition} \\
    \midrule
    \multicolumn{3}{@{}l}{\textit{Overall}} \\
    Samples & 1{,}075 & Number of image pairs. \\
    Overall diffs. & 6{,}713 & Total verified difference items. \\
    Avg. diffs. & 6.25 & Mean number of differences per pair. \\
    \midrule
    \multicolumn{3}{@{}l}{\textit{\textbf{Difference Type}}} \\
    Object & 2{,}314 & Object-level additions, deletions, or replacements. \\
    Attribute & 1{,}760 & Attribute modifications (e.g., size, color). \\
    Action \& Pose & 1{,}153 & Changes in actions or body poses. \\
    Spatial & 1{,}002 & Spatial rearrangements (e.g., moving objects). \\
    Composition & 834 & Compositional differences, including cropping. \\
    Text \& Symbol & 831 & Addition, removal, or modification of text/symbols. \\
    Background & 806 & Alterations to the scene background. \\
    Camera/View & 571 & Camera movement or viewpoint changes. \\
    Image Prop. & 425 & Global changes (e.g., resolution, saturation). \\
    Expression & 333 & Changes in facial expressions. \\
    \bottomrule
  \end{tabularx}
  \caption{DiffCap-Bench statistics and category definitions.}
  \label{tab:ierplus-stats}
\end{table}

\begin{table*}[t]
\centering
\scriptsize
\setlength{\tabcolsep}{6pt}
\begin{tabular*}{\textwidth}{@{\extracolsep{\fill}} l l c c c c c c @{}}
\toprule
Dataset & Source & Avg.\ Chg. & Avg.\ Wd. & Human & Eval. & Metrics & Hall. \\
\midrule
CLEVR-Change~\cite{clevrchange}   & 3D render      & 1    & 8    & \xmark & Lexical & B/M/R/C  & \xmark \\
Spot-the-Diff~\cite{spotdiff}   & Surv.\ video   & 1.86    & 19   & \cmark & Lexical & B/M/R/C  & \xmark \\
Image-Edit-Request~\cite{imgeditrequest}          & Diverse real   & 1    & 8    & \cmark & Lexical & B/M/R/C  & \xmark \\
OneDiff~\cite{onediff}        & Syn.\ \& real  & 1    & 15   & \xmark & Lexical & B/M/R/C  & \xmark \\
DiffTell~\cite{difftell}       & Syn.\ \& real  & 1    & 9.72 & \cmark & Lexical & B/M/R/C  & \xmark \\
OmniDiff~\cite{omnidiff}       & 3D \& real     & 2.5  & 60   & \cmark & Lexical & B/M/R/C  & \xmark \\
\midrule
DiffCap-Bench (Ours) & 2D, 3D \& real & 6.25 & 175  & \cmark & LLM-as-a-Judge & R$^{*}$/P$^{*}$/$F_{1}^{*}$/H & \cmark \\
\bottomrule
\end{tabular*}
\caption{Comparison of DiffCap-Bench with existing IDC benchmarks (Surv.: Surveillance, Syn.: Synthetic, Chg.: Changes, Wd.: Words; B/M/R/C: BLEU, METEOR, ROUGE-L, CIDEr; R$^{*}$/P$^{*}$/$F_{1}^{*}$/H: Recall$^{*}$, Precision$^{*}$, $F_{1}^{*}$, Hallucination Rate).}
\label{tables:BenchmarkCompare}
\end{table*}

\input{sections/DiffCapBench}

\begin{table*}[t]
  \centering
  \scriptsize
  \setlength{\tabcolsep}{5pt}
  \begin{tabular*}{\textwidth}{@{\extracolsep{\fill}} l *{10}{c} *{4}{>{\columncolor{gray!20}}c} @{}}
  \toprule
  Model & \rotatebox[origin=c]{45}{Comp.} & \rotatebox[origin=c]{45}{Back} & \rotatebox[origin=c]{45}{Attr.} & \rotatebox[origin=c]{45}{Text} & \rotatebox[origin=c]{45}{Cam.} & \rotatebox[origin=c]{45}{Expr.} & \rotatebox[origin=c]{45}{Spat.} & \rotatebox[origin=c]{45}{Img.} & \rotatebox[origin=c]{45}{Obj.} & \rotatebox[origin=c]{45}{Act.} & \rotatebox[origin=c]{45}{R$^{*}$} & \rotatebox[origin=c]{45}{P$^{*}$} & \rotatebox[origin=c]{45}{$F_{1}^{*}$} & \rotatebox[origin=c]{45}{H} \\
  \midrule
  \multicolumn{15}{c}{\textbf{Commercial}} \\
  \midrule
  Gemini 3 Pro~\cite{gemini}     & \underline{88.0} & \underline{89.4} & \underline{80.8} & 88.0 & \underline{86.5} & \underline{72.1} & \textbf{83.5} & \underline{87.8} & \textbf{87.7} & \underline{75.5} & \textbf{76.9} & \textbf{93.6} & \textbf{83.1} & \textbf{2.6} \\
  GPT 5.2~\cite{gpt5}    & \textbf{91.5} & 89.0 & 77.1 & 86.3 & \textbf{89.4} & 30.0 & \underline{82.7} & \textbf{89.1} & 83.0 & 62.9 & 70.5 & \underline{85.2} & \underline{75.1} & 12.0 \\
  Seed1.6-VL~\cite{seedvl}  & 68.0 & 87.2 & 77.6 & 85.7 & 59.3 & 44.3 & 70.5 & 83.3 & 83.3 & 63.9 & 67.4 & 78.2 & 69.7 & 23.9 \\
  Seed1.6-VL-Thinking~\cite{seedvl}  & 73.6 & 87.8 & 79.2 & 86.3 & 65.3 & 55.9 & 75.6 & 83.6 & \underline{86.2} & 67.5 & 70.5 & 80.4 & 73.1 & 20.9 \\
  Qwen3VL-Plus~\cite{qwen3vl}     & 79.3 & 87.9 & \textbf{82.4} & \textbf{90.5} & 75.1 & \textbf{79.1} & 78.6 & 85.9 & 85.3 & \textbf{75.8} & 71.3 & 83.3 & 74.9 & 18.3 \\
  Kimi 2.5~\cite{kimi2d5} & 77.5 & \textbf{90.0} & 80.0 & 87.5 & 71.1 & 48.3 & 75.0 & 87.2 & 85.7 & 66.4 & 71.8 & 76.0 & 71.3 & 37.7 \\
  Kimi 2.5 Thinking~\cite{kimi2d5} & 77.4 & 86.9 & 80.0 & 88.0 & 74.5 & 60.6 & 78.0 & 82.1 & 86.6 & 66.8 & \underline{71.9} & 81.6 & 74.5 & 18.9 \\
  Grok 4.2 Beta & 75.8 & 82.7 & 70.3 & 76.5 & 72.7 & 71.6 & 73.8 & 72.7 & 77.4 & 66.8 & 63.5 & 81.1 & 69.6 & \underline{10.5} \\
  \midrule
  \multicolumn{15}{c}{\textbf{Open-Source}} \\
  \midrule
  InternVL3.5-8B~\cite{internvl3d5}  & 62.9 & 71.9 & 62.0 & 78.9 & 64.9 & 58.2 & 63.0 & 77.2 & 69.3 & 59.9 & 51.1 & 60.0 & 51.9 & 62.6 \\
  InternVL3.5-38B~\cite{internvl3d5} & 59.2 & 73.8 & 63.5 & 78.5 & 57.2 & 63.8 & 63.1 & 78.9 & 70.8 & 62.7 & 53.3 & 65.6 & 55.5 & 51.8 \\
  Qwen3VL-8B-Instruct~\cite{qwen3vl}    & 69.2 & 77.3 & 70.4 & 77.3 & 70.3 & 63.7 & 67.0 & 76.2 & 72.4 & 66.9 & 56.4 & 74.7 & 61.7 & 26.1 \\
  Qwen3VL-8B-Thinking~\cite{qwen3vl}    & 66.7 & 81.4 & 73.3 & 83.7 & 62.9 & 48.3 & 71.1 & 74.1 & 79.4 & 63.3 & 60.4 & 77.0 & 65.5 & 20.2 \\
  Qwen3VL-32B-Instruct~\cite{qwen3vl}   & 72.3 & 81.3 & 68.8 & 76.2 & 71.7 & 66.2 & 71.5 & 82.9 & 71.9 & 65.2 & 62.7 & 67.7 & 63.8 & 31.3 \\
  Qwen3VL-32B-Thinking~\cite{qwen3vl}  & 69.2 & 85.7 & 78.6 & \underline{88.1} & 63.3 & 54.2 & 75.0 & 75.9 & 85.0 & 68.0 & 65.1 & 78.7 & 68.7 & 25.8 \\
  Step3-VL-10B~\cite{step3vl} & 77.2 & 84.1 & 70.9 & 80.9 & 77.5 & 37.8 & 73.9 & 75.7 & 79.7 & 57.7 & 62.3 & 76.1 & 66.1 & 22.8 \\
  MiMo-VL-7B-RL-2508~\cite{mimovl}  & 75.2 & 83.0 & 74.9 & 84.3 & 72.7 & 45.0 & 73.9 & 75.8 & 81.4 & 63.5 & 62.3 & 74.8 & 64.7 & 34.0 \\
  \bottomrule
  \end{tabular*}
  \caption{Performance of MLLMs on DiffCap-Bench. Comp., Back., Attr., Text, Cam., Expr., Spat., Img., Obj., and Act. denote Composition, Background, Attribute, Text \& Symbol, Camera \& Viewpoint, Expression, Spatial, Image Property, Object, and Action \& Pose, respectively. Category columns report macro-average $F_{1}^{*}$, while other columns report sample-level average $F_{1}^{*}$.}
  \label{tables:MainResult}
  \end{table*}

\begin{figure*}[t]
\centering
\includegraphics[width=0.82\textwidth]{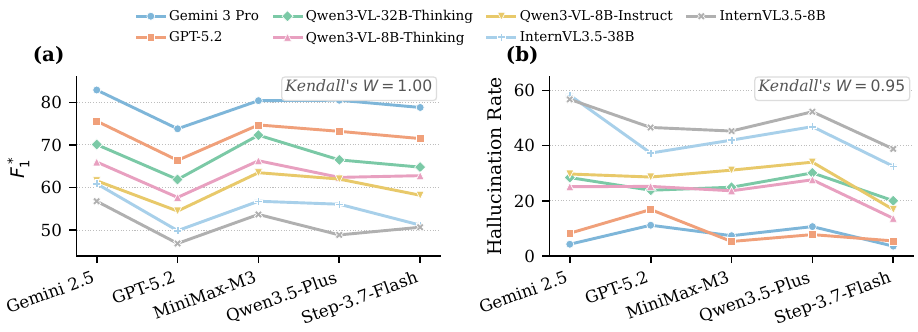}
\caption{Cross-judge ranking consistency on DiffCap-Bench. We re-evaluate seven representative MLLMs under five independent judges and report (a)~$F_{1}^{*}$ and (b)~Hallucination Rate. Absolute scores vary across judges, but relative rankings remain highly stable (Kendall's $W{=}1.00$ for $F_{1}^{*}$ and $W{=}0.95$ for Hallucination Rate).}
\label{fig:cross_judge_rank}
\end{figure*}

\begin{figure*}[t]
  \centering
  \includegraphics[width=0.9\textwidth]{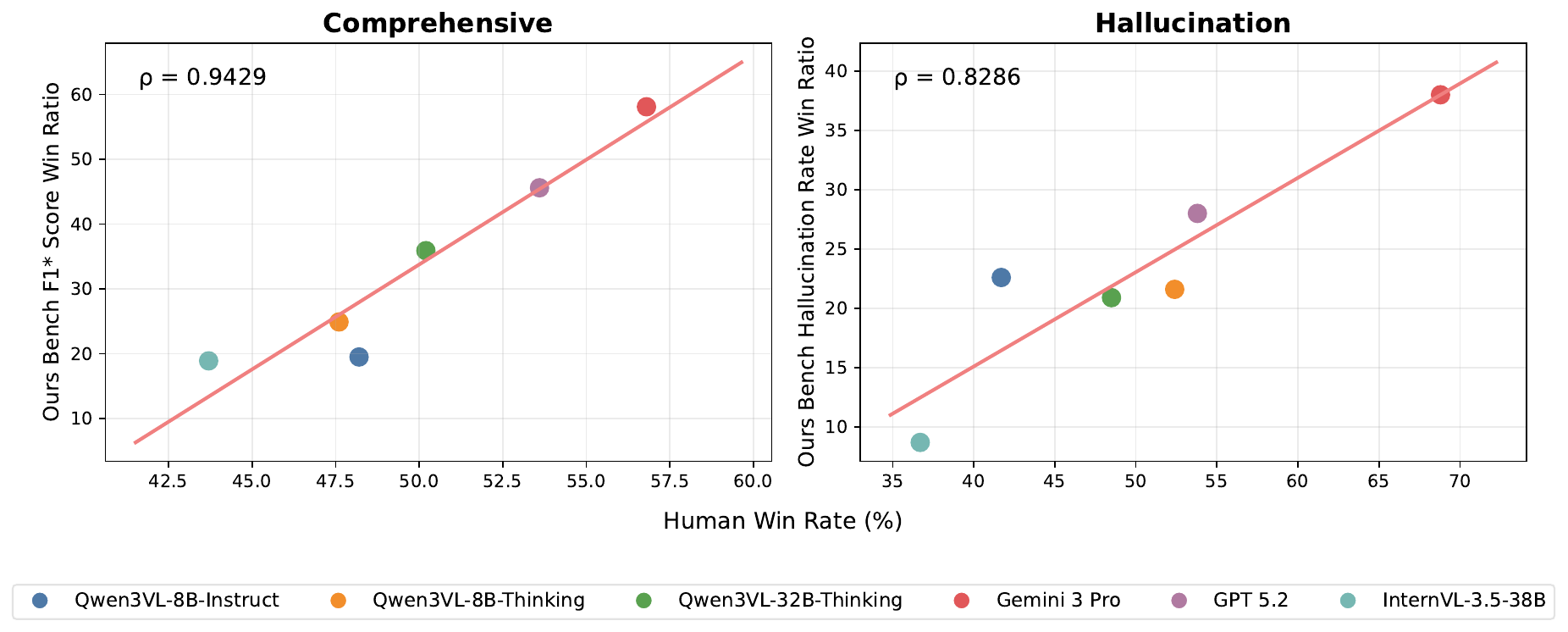}
  \caption{Correlation between DiffCap-Bench metrics and human expert judgments across six representative MLLMs. Per-model win rates are computed under both evaluation protocols, and rank consistency is measured via Spearman's $\rho$, yielding $\rho=0.94$ for $F_{1}^{*}$ and $\rho=0.82$ for Hallucination Rate.}
  \label{fig:human_correlation}
\end{figure*}

\section{Experiment}
\label{sec:Experiment}
\input{sections/experiment}

\section{Effectiveness Validation}
\label{sec:Effectiveness Provement}

\input{sections/EffectivenessProvement}

\section{Conclusion}
\label{sec:conclusion}

\input{sections/conclusion}

\bibliography{diffcap}

\clearpage
\appendix
\input{appendix/appendix}

\end{document}

%% file: sections/abstract.tex
Image Difference Captioning (IDC) generates natural language descriptions that precisely identify differences between two images, serving as a key task for fine-grained change perception, cross-modal reasoning, and image editing data construction. However, existing benchmarks lack data diversity and compositional complexity, while their evaluation protocols rely on lexical-overlap metrics (e.g., BLEU, METEOR) that fail to capture semantic consistency or penalize hallucinations, preventing a comprehensive and robust evaluation of multimodal large language models (MLLMs) on IDC. To address these gaps, we introduce \textbf{DiffCap-Bench}, a comprehensive IDC benchmark covering ten difference categories. Furthermore, we propose an LLM-as-a-Judge evaluation protocol grounded in human-validated \textit{Difference Lists}, enabling a robust assessment of models' ability to capture and describe visual changes. Through extensive evaluation of state-of-the-art MLLMs, we reveal significant gaps between proprietary and open-source models and the critical role of reasoning capability. Further analysis shows that our framework demonstrates stable cross-judge rankings, strong alignment with human expert judgments, and strong correlation with downstream image editing data construction quality. These findings establish DiffCap-Bench as both a reliable IDC evaluation framework and a practical predictor of downstream utility.

%% file: sections/introduction.tex
Image Difference Captioning (IDC) provides precise natural language descriptions of visual discrepancies between image pairs, serving as a foundational task for change detection and visual reasoning~\cite{showandtell,vlmsubtle}. In particular, IDC also serves as a critical upstream component for image editing pipelines by identifying differences between two images to provide intermediate edit instructions whose accuracy directly determines the quality of the resulting image editing data~\cite{hqedit,imgedit}. With the rapid advancement of Multimodal Large Language Models (MLLMs) and their demonstrated strong performance on open-ended vision-language tasks such as image captioning~\cite{caprl,omnicaptioner}, how well these models perform on IDC remains systematically unexplored. This gap stems primarily from the limitations of existing benchmarks that fall short on both the data side and the evaluation side, rendering them unable to provide a robust and reliable assessment of MLLMs' IDC capabilities.

On the data side, existing IDC benchmarks remain limited in two key aspects: they are neither comprehensive enough in difference coverage nor challenging enough in compositional complexity. Earlier datasets exhibit restricted diversity in different ways: Spot-the-Diff~\cite{spotdiff} is mainly built from surveillance-style videos with constrained scenarios, while CLEVR-Change~\cite{clevrchange} focuses on synthetic settings; both typically contain few differences per pair. More recent datasets, including Image-Edit-Request~\cite{imgeditrequest}, OneDiff~\cite{onediff}, DiffTell~\cite{difftell}, and OmniDiff~\cite{omnidiff}, improve realism, yet still often exhibit limited compositionality, incomplete coverage, or annotation noise. As a result, they underrepresent the breadth and difficulty of IDC in modern image editing contexts, where multiple coupled differences can co-occur across objects, attributes, text, viewpoint, and global image properties.

\begin{figure}[t]
  \centering
  \includegraphics[width=0.82\columnwidth]{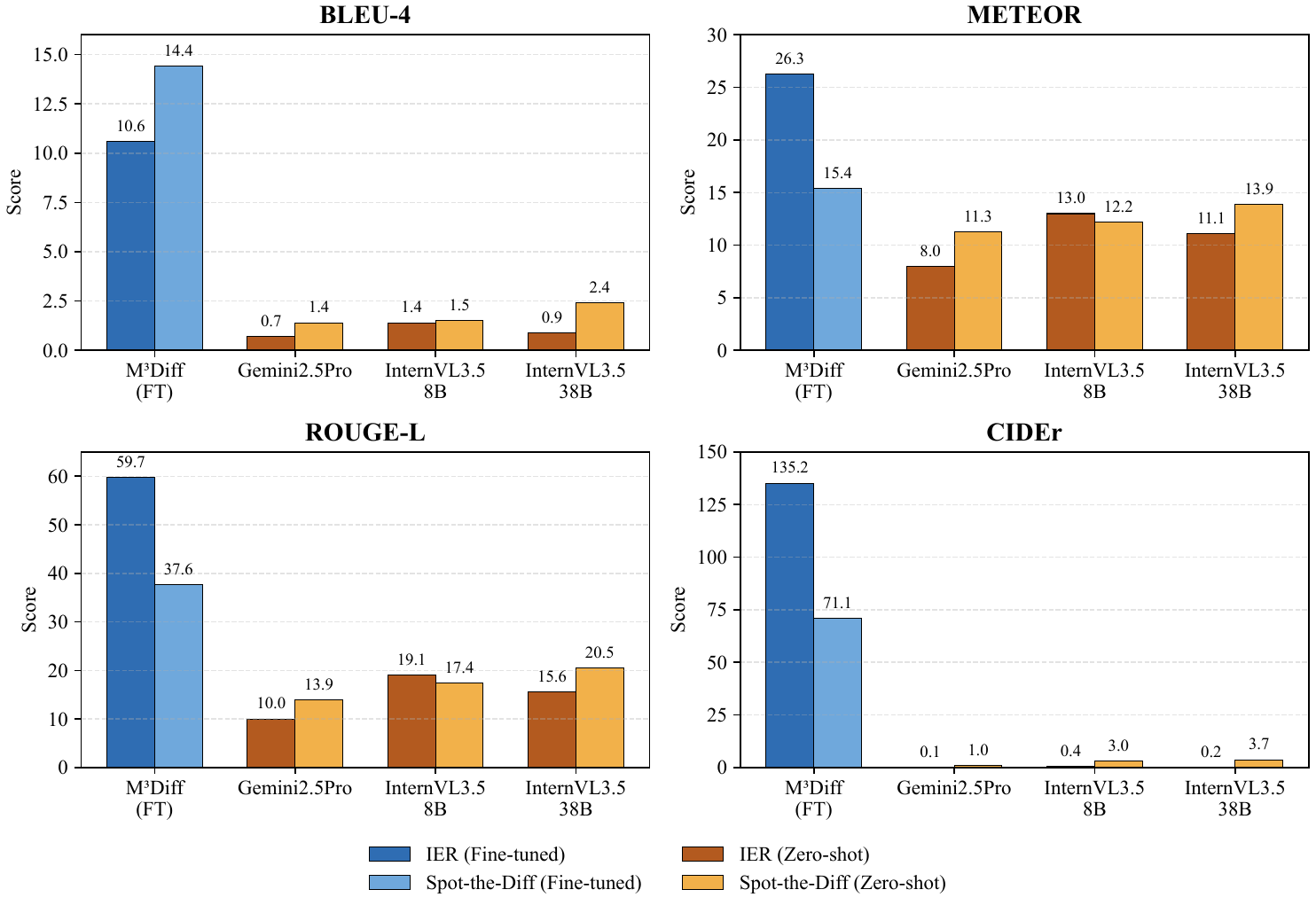}
  \caption{Under lexical-overlap metrics, zero-shot MLLMs score far below fine-tuned model on the traditional IDC benchmarks (Image-Edit-Request, Spot-the-Diff).}
  \label{fig:idc_benchmark_results}
\end{figure}

On the evaluation side, most previous IDC benchmarks~\cite{onediff,omnidiff,spotdiff} rely on lexical-overlap metrics (e.g., BLEU~\cite{bleu}, METEOR~\cite{meteor}, ROUGE-L~\cite{rouge}, CIDEr~\cite{cider}), which are brittle for open-form difference descriptions~\cite{metricproblem,metricproblem2}. As shown in Figure~\ref{fig:idc_benchmark_results}, zero-shot scores of strong MLLMs remain uniformly low on traditional IDC benchmarks (Image-Edit-Request~\cite{imgeditrequest} and Spot-the-Diff~\cite{spotdiff}) under these metrics, falling far below the dataset-fine-tuned $\mathrm{M}^{3}\mathrm{Diff}$~\cite{omnidiff} and showing poor separability across models. This is largely because free-form MLLM outputs, which often differ in language style and response format, are poorly matched to n-gram lexical overlap. Moreover, these metrics cannot reliably capture semantic correctness and fail to penalize hallucinations of non-existent differences. Consequently, they cannot provide a reliable and robust evaluation of modern MLLMs on IDC.

To address these issues, we introduce \textbf{DiffCap-Bench}, a comprehensive and challenging benchmark for realistic and compositional IDC in modern image editing contexts. DiffCap-Bench is designed with two principles: coverage of diverse real image editing scenarios and semantic reliability of evaluation. For data construction, we combine diverse real-world sampling (e.g., web images, videos, advertisements, posters) with controlled synthetic generation (2D compositing and 3D rendering), yielding 1,075 high-quality image pairs that systematically cover ten difference categories.

Beyond data diversity and challenge, we propose an \emph{LLM-as-a-Judge}~\cite{llmAsAJudgeSurvey} evaluation protocol grounded in human-validated \textit{Difference Lists} that enables robust assessment of models' ability to capture and describe visual changes. Rather than comparing generated text against a single reference sentence, our protocol operates on human-validated atomic difference items and evaluates completeness and faithfulness through Recall$^{*}$, Precision$^{*}$, $F_{1}^{*}$, and Hallucination Rate. This design enables robust assessment under linguistic variation and captures whether a model identifies correct differences while avoiding unsupported claims.

Leveraging DiffCap-Bench, we systematically evaluate 16 proprietary and open-source MLLMs and find: (1) a clear performance gap between proprietary and open-source models; (2) the superiority of \emph{Thinking} variants over their base counterparts, highlighting the role of reasoning in accuracy and faithfulness; and (3) a bottleneck on action, pose, and expression changes, versus stronger performance on background, text, and image-property differences.

To assess reliability, we analyze DiffCap-Bench from three perspectives---cross-judge ranking consistency, alignment with human judgment, and downstream utility. We find highly stable rankings across independent judges, strong agreement between our automatic metrics and human experts, and clear correlation between benchmark scores and downstream image-editing quality under model-generated relabeling data. These findings confirm DiffCap-Bench as a robust IDC evaluation framework and a reliable proxy for real-world utility.

In summary, our contributions are threefold:
\begin{itemize}
\item We introduce \textbf{DiffCap-Bench}, a comprehensive, challenging and robust benchmark designed for evaluation of MLLMs in fine-grained image difference captioning.
\item We propose a semantically grounded evaluation framework based on human-validated Difference Lists, and to our knowledge provide the first systematic evaluation of MLLMs on IDC, covering 16 state-of-the-art models and revealing key limitations of current systems.
\item We validate DiffCap-Bench via cross-judge consistency, human alignment, and downstream correlation, showing it as a robust evaluation protocol and a proxy for MLLM utility in automated image-editing data engineering.
\end{itemize}

%% file: sections/relatedworks.tex
\subsection{Image Difference Captioning Benchmarks.} Early IDC benchmarks mostly focus on constrained scenarios and simple difference types. Spot-the-Diff~\cite{spotdiff} is built from surveillance videos and shows limited variation patterns, making it less suited to evaluating model robustness in diverse real-world settings. CLEVR-Change~\cite{clevrchange} and CLEVR-DC~\cite{clevrdc} rely on synthetic desktop objects and mainly cover basic operations such as object addition, deletion, and replacement.
More recent datasets improve realism and scale, but key limits remain. Image-Edit-Request~\cite{imgeditrequest} introduces natural images with textual editing requests, yet its difference descriptions are typically short and rarely involve compositional or multi-step changes. OneDiff~\cite{onediff} combines real and synthetic images, but suffers from annotation noise and uneven coverage of key difference types, such as text edits and OCR-sensitive changes. DiffTell~\cite{difftell} covers a wider range of differences (e.g., background, local object edits, text, and style), but most samples still center on single-type changes and do not fully capture complex, multi-aspect differences. OmniDiff~\cite{omnidiff} integrates real scenes with 3D simulation to increase diversity, yet still underrepresents camera-viewpoint and compositional shifts. Overall, existing datasets remain insufficient in both broad coverage and compositional difficulty.

\subsection{Evaluation Methods for IDC.} 

Standard IDC evaluation relies on lexical-overlap metrics such as BLEU~\cite{bleu}, CIDEr~\cite{cider}, METEOR~\cite{meteor}, and ROUGE-L~\cite{rouge}. However, these lexical-level measures are brittle for open-form descriptions~\cite{metricproblem,metricproblem2}, as they favor surface similarity over semantic fidelity. As a result, they often fail to detect key factual errors or penalize hallucinations; for example, semantically opposite phrases like \textit{raised left hand} and \textit{raised right hand} may still receive similar overlap scores. To address these limits, recent studies on LLM-as-a-Judge~\cite{llmAsAJudgeSurvey} have shown better alignment with human judgment in open-ended tasks, indicating their potential for robust evaluation.


%% file: sections/DiffCapBench.tex
To address the limitations of insufficiently comprehensive and challenging data and the lack of robust evaluation in IDC, we introduce \textbf{DiffCap-Bench} as illustrated in Fig.~\ref{fig:pipeline}. In the following subsections, we first describe diverse image-pair construction and difference list labeling. We then report comprehensive dataset statistics and finally compare DiffCap-Bench with existing IDC benchmarks.

\subsection{Image Pair Construction}
\label{sec:pair-construction}
To ensure data diversity and broad coverage of realistic difference types, we construct data through both sampling and synthesis pipelines, with quality filtering applied to ensure the reliability of the resulting image pairs.

\noindent\textbf{Multisource Sampling}
We collect image pairs from multiple sources to maximize coverage of real-world differences, including web images, video frames, advertisements, and posters. These sources are rich in human-centric content and naturally contain diverse, realistic transformations. From this pool, we identify candidate image pairs for further filtering.

\noindent\textbf{Synthetic Image Pairs via 2D Compositing and 3D Rendering}
We generate synthetic image pairs to further improve dataset comprehensiveness and challenge through controlled compositional editing. For 2D synthesis, we use Canva, an online graphic design and layout platform, to perform layered compositing and element-level modifications that induce diverse visual discrepancies, ranging from object manipulation and attribute shifts to background and text alterations. For 3D synthesis, we use Blender, an open-source 3D creation suite, to render scenes with systematically varied elements, enabling structured and controllable difference patterns.

\noindent\textbf{Quality Filtering}
To ensure both data quality and benchmark challenge, we apply strict filtering to all collected pairs. We employ a strong MLLM, Qwen3VL-Plus~\cite{qwen3vl}, as the automatic filter with three criteria: (1) \emph{Difference Existence}: the two images must contain real differences rather than being identical; (2) \emph{Visual Similarity}: the pair must remain semantically related rather than depicting unrelated scenes; and (3) \emph{Image Quality}: both images must satisfy quality requirements (e.g., resolution and clarity). All retained pairs are verified by human experts, and pairs deemed insufficiently challenging are removed. Following this pipeline, we curate a final set of \textit{1,075} high-quality image pairs. The MLLM filtering prompts, detailed criteria, and the challengingness filtering procedure are provided in Appendix.

\subsection{Difference List Labeling}
\label{sec:diff-list}
To enable robust evaluation, we construct the human-validated \emph{Difference List} comprising atomic difference items for each image pair through a three-stage pipeline in Fig.~\ref{fig:pipeline}. First, multiple MLLMs (Qwen3VL-30B-A3B-Instruct~\cite{qwen3vl}, Gemini 2.5 Pro~\cite{gemini}, and Gemini 3 Pro~\cite{gemini}) are prompted to describe pairwise differences, and the resulting descriptions are segmented into atomic units and merged into a candidate pool (typically 10--30 items per pair) to improve recall. Second, we employ a professional annotation team with extensive image-caption experience and conduct a 10-workday campaign with task-specific training. Each candidate is reviewed in a double-blind manner by at least two annotators and labeled as: (1) \emph{correct}: the difference is real and accurately described; (2) \emph{incorrect-but-fixable}: the difference is real but described inaccurately and requires correction; (3) \emph{hallucinated}: the difference does not exist in the image pair; or (4) \emph{indistinguishable}: the descriptions cannot be reliably determined. Annotators may also add missing differences; in the first round, about 32\% of candidates required correction or removal. Finally, all valid differences, including correct, expert-corrected, and expert-added items, are aggregated into a \textbf{Key Difference List}, while indistinguishable ones are retained as \textbf{Indistinguishable Items}, constituting the human-validated \emph{Difference List} for each pair.

\subsection{Dataset Statistics}
\label{sec:dataset-stats}
As shown in Table~\ref{tab:ierplus-stats}, we collect 1,075 image pairs, and after expert annotation, the dataset contains 6,713 difference items, with an average of 6.25 differences per pair and 175 words per \emph{Difference List}. To analyze difference-type distribution, we categorize each difference into predefined edit categories using an LLM-assisted classification process (see Appendix), allowing multi-label assignments as a single difference can involve multiple aspects. All annotations are verified by humans.

\subsection{Comparisons with Existing Benchmarks}
\label{sec:benchmark-compare}
We compare DiffCap-Bench with representative Image Difference Captioning benchmarks in terms of data source, difference complexity, annotation protocol, and evaluation methodology. As shown in Table~\ref{tables:BenchmarkCompare}, existing datasets typically contain only a limited number of differences per image pair and rely on lexical-overlap metrics. Moreover, hallucination detection is not explicitly modeled. In contrast, DiffCap-Bench contains more differences per image pair (6.25 on average) and longer descriptions, reflecting higher compositional complexity and benchmark challenge. Furthermore, we adopt a Difference List-based \emph{LLM-as-a-Judge} framework that measures completeness and faithfulness through Recall$^{*}$, Precision$^{*}$, $F_{1}^{*}$, and Hallucination Rate, enabling a more robust and discriminative evaluation mechanism.

%% file: sections/experiment.tex
We conduct extensive experiments to evaluate the performance of state-of-the-art MLLMs on our benchmark.
This section presents our experimental pipeline and corresponding key findings, which are organized as follows: evaluation protocol and metrics, evaluated models and the results.

\subsection{Evaluation Protocol and Metrics}\label{sec:evaluation}

To provide a robust evaluation of generated difference descriptions, we propose a bidirectional protocol based on the predefined \textit{Difference List}. For a given image pair, the model generates a predicted description $a_1$ as follows:
\begin{equation}
a_1 = \mathcal{P}(y \mid \text{image pair}, \text{prompt})
\end{equation}

The evaluation protocol then prompts an LLM-as-a-Judge to apply a two-stage checking process to ensure comprehensive coverage and factual faithfulness: \textbf{Forward Check (Coverage \& Precision)} measures coverage and factual correctness with respect to the ground truth. Each item in the \textit{Key Difference List} is cross-referenced with $a_1$ to determine whether it is captured and correctly described. We define $N_{hr}$ as the number of correctly identified differences and $N_{hw}$ as the number of differences that are mentioned but incorrectly described. \textbf{Backward Check (Redundancy \& Hallucination)} identifies redundant or erroneous content in $a_1$. Any predicted element not found in the \textit{Key Difference List} is cross-referenced with the \textit{Indistinguishable Items}. Matches in the latter are treated as ambiguous and excluded from penalties, as even expert annotators cannot provide reliable determinations for these cases. The remaining extra elements are then classified as either valid elaborations or hallucinations, with $N_{hallu}$ denoting the number of identified hallucinations. Based on these statistics, we define the following metrics:
\paragraph{Recall$^{*}$} Quantifies the model's capacity to encompass all ground-truth differences:
\begin{equation}
\text{Recall}^{*} = \frac{N_{hr}}{|L_{key}|}
\end{equation}
where $|L_{key}|$ denotes the total number of difference items in the \textit{Key Difference List}.

\paragraph{Precision$^{*}$} Measures the proportion of correct descriptions among all predicted differences:
\begin{equation}
\text{Precision}^{*} = \frac{N_{hr}}{N_{hr} + N_{hw} + N_{hallu}}
\end{equation}

\paragraph{$F_{1}^{*}$ Score} The harmonic mean of Recall$^{*}$ and Precision$^{*}$, serving as a \textbf{holistic metric} to balance coverage and factual integrity:
\begin{equation}
F_{1}^{*} = \frac{2 \cdot \text{Recall}^{*} \cdot \text{Precision}^{*}}{\text{Recall}^{*} + \text{Precision}^{*}}
\end{equation}

\paragraph{Hallucination Rate} Quantifies the frequency of hallucinations relative to the difference density of the scene:
\begin{equation}
\text{Hallucination Rate} = \frac{N_{hallu}}{|L_{key}|}
\end{equation}

By normalizing hallucinations by the number of ground-truth differences, this metric imposes a stronger penalty on spurious predictions in simpler scenes with fewer visual discrepancies. The prompt for the judge LLM and a representative evaluation case study are provided in the Appendix.

\subsection{Evaluated Models}
We conduct a thorough evaluation across 16 state-of-the-art MLLMs, including both commercial and open-source models.
Commercial models include Gemini 3 Pro~\cite{gemini}, GPT 5.2~\cite{gpt5}, Seed1.6-VL~\cite{seedvl}, Seed1.6-VL-Thinking~\cite{seedvl}, Qwen3VL-Plus~\cite{qwen3vl}, Kimi 2.5~\cite{kimi2d5}, Kimi 2.5 Thinking~\cite{kimi2d5}, and Grok 4.2 Beta. Open-source models include InternVL3.5 (8B/38B)~\cite{internvl3d5}, Step3-VL-10B~\cite{step3vl}, MiMo-VL-7B-RL-2508~\cite{mimovl}, Qwen3VL-Instruct (8B/32B)~\cite{qwen3vl}, and Qwen3VL-Thinking (8B/32B)~\cite{qwen3vl}, ranging from 7B to 38B parameters.

For all models, we use the same system prompt and apply the evaluation protocol in section Evaluation Protocol and Metrics, with Gemini 2.5 Pro~\cite{gemini} serving as the judge model. The system prompt and inference configuration for all evaluated models are provided in the Appendix.

\subsection{Evaluation Results of DiffCap-Bench}
Table~\ref{tables:MainResult} reports results on DiffCap-Bench, from which we draw the following observations.

\noindent\textbf{A clear performance gap remains between commercial and open-source MLLMs.} Among commercial models, Gemini 3 Pro achieves state-of-the-art performance with an $F_{1}^{*}$ of 83.1\%, substantially surpassing GPT 5.2 (75.1\%) and the leading open-source model, Qwen3VL-32B-Thinking (68.7\%). While commercial models generally maintain $F_1^{*}$ above 70\%, most open-source models struggle in the 60--70\% range. This gap is clearer in Hallucination Rate: Gemini 3 Pro reaches only 2.6\%, whereas other models typically exhibit 20\%--30\%, and the InternVL3.5 series exceeds 50\%, indicating a major challenge in factual faithfulness on IDC.

\noindent\textbf{Reasoning ability and Chain-of-Thought significantly enhance IDC performance.} Across identical architectures, \emph{Thinking} variants consistently outperform their base counterparts. Seed1.6-VL-Thinking improves $F_{1}^{*}$ by 3.4\% and reduces Hallucination Rate by 3\%; Qwen3VL-8B-Thinking gains 3.8\% $F_{1}^{*}$ and cuts hallucinations by 6\% over Instruct; Qwen3VL-32B-Thinking further yields a 5\% $F_{1}^{*}$ gain and a 5.5\% hallucination drop. Most strikingly, Kimi 2.5 Thinking reduces hallucinations by 19\% with a modest $F_{1}^{*}$ gain (74.5\% vs.\ 71.3\%). These results suggest that CoT reasoning is pivotal for perceiving and describing subtle visual discrepancies. Case studies are provided in the Appendix.


\noindent\textbf{Existing models excel at perceiving background, text, and image property changes, yet struggle with action, pose, and expression discrepancies.} All open-source models yield macro-average $F_1^{*}$ below 70\% on ``action \& pose'' and ``expression''. Expression is the hardest: except for Gemini 3 Pro and Qwen3VL-Plus, even leading commercial models perform poorly---e.g., GPT 5.2 reaches only 30.0\%, Seed1.6-VL variants stay below 60\%, and other open-source models range from 30\% to 66\%. These results reveal a bottleneck in modeling non-rigid or nuanced visual discrepancies.

%% file: sections/EffectivenessProvement.tex
\begin{table*}[t]
    \centering
    \scriptsize
    \setlength{\tabcolsep}{5pt}
    \renewcommand{\arraystretch}{0.88}
    \begin{tabular*}{0.85\textwidth}{@{\extracolsep{\fill}} l l c c c c c c c @{}}
    \toprule
    \multirow{2}{*}{Base Model} & \multirow{2}{*}{Relabeling LVLM} & \multicolumn{5}{c}{ImgEdit Eval} & \multicolumn{2}{c}{DiffCap-Bench} \\
    \cmidrule(lr){3-7} \cmidrule(lr){8-9}
     & & IC & VS & PC & Overall & Rank & F1$^*$ & Rank \\
    \midrule
    \multirow{5}{*}{FLUX.1-Kontext-dev} 
     & InternVL3.5-8B      & 2.15 & 2.08 & 2.11 & 2.11 & 5 & 51.9 & 5 \\
     & InternVL3.5-38B     & 2.27 & 2.05 & 2.15 & 2.15 & 4 & 55.5 & 4 \\
     & Qwen3VL-8B-Instruct & 2.32 & 2.23 & 2.08 & 2.21 & 2 & 61.7 & 3 \\
     & Qwen3VL-8B-Thinking & 2.33 & 2.03 & 2.12 & 2.16 & 3 & 65.5 & 2 \\
     & Qwen3VL-32B-Thinking& 3.12 & 2.95 & 2.87 & 2.98 & 1 & 68.7 & 1 \\
    \bottomrule
    \end{tabular*}
    \caption{Downstream image editing performance of FLUX.1-Kontext-dev fine-tuned on datasets relabeled by different MLLMs. IC, VS, and PC denote Instruction Compliance, Visual Seamlessness, and Physical Consistency, which are ImgEdit scoring dimensions.}
    \label{tables:downstream}
\end{table*}

We then validate the effectiveness of DiffCap-Bench from three perspectives: cross-judge robustness, human alignment, and correlation with image-editing data production.

\subsection{Cross-Judge Ranking Consistency}
\label{sec:cross-judge}
To test sensitivity to the choice of judge, we randomly sample 200 items and re-evaluate seven representative MLLMs under five independent judges. As shown in Figure~\ref{fig:cross_judge_rank}, absolute $F_{1}^{*}$ and Hallucination Rate vary across judges, but rankings remain highly stable (Kendall's $W{=}1.00$ for $F_{1}^{*}$; $W{=}0.95$ for Hallucination Rate), indicating that conclusions are anchored by the human-validated Difference List rather than a particular judge, and further demonstrating the robustness of DiffCap-Bench.

\subsection{Alignment with Human Judgment}
We sample 100 instances and evaluate six MLLMs (Gemini 3 Pro, GPT 5.2, Qwen3VL-32B-Thinking, Qwen3VL-8B-Instruct, Qwen3VL-8B-Thinking, and InternVL3.5-38B) under our automatic metrics and human assessment. Experts score holistic quality and hallucination severity (1--5), from which we compute per-model win rates. As shown in Figure~\ref{fig:human_correlation}, automatic metrics align with human judgments (Spearman's $\rho{=}0.9429$ for $F_{1}^{*}$; $\rho{=}0.8286$ for Hallucination Rate), confirming a reliable, human-aligned protocol for IDC. Win-rate details and scoring rubrics are provided in the Appendix.

\subsection{Correlation with Data Production}
IDC is a key upstream step for constructing image-editing data. To assess whether DiffCap-Bench predicts downstream utility, we use five MLLMs (InternVL3.5-8B/38B, Qwen3VL-8B-Instruct/Thinking, and Qwen3VL-32B-Thinking) to separately relabel 300K pairs from Pico-Banana-400k~\cite{picobanana400k} and NanoBanana-150k~\cite{nanobanana150k}, yielding five training datasets with identical image pairs but different difference captions. The base model \texttt{FLUX.1-Kontext-dev}~\cite{flux} is then fine-tuned separately on each relabeled set under the same training setting, and evaluated on 500 held-out pairs following the ImgEdit~\cite{imgedit} evaluation protocol. The training details are provided in the Appendix.

As shown in Table~\ref{tables:downstream}, DiffCap-Bench $F_{1}^{*}$ rankings align closely with Overall editing scores (Spearman's $\rho{=}0.90$), and Instruction Compliance (IC) follows a notably consistent trend with $F_{1}^{*}$. In particular, Qwen3VL-32B-Thinking attains the best Overall (2.98) and $F_{1}^{*}$ (68.7\%). These results show DiffCap-Bench is highly correlated with downstream utility and serves as a proxy for MLLM utility in automated image-editing data engineering.

%% file: sections/conclusion.tex
In this work, we present \textbf{DiffCap-Bench}, a comprehensive, challenging, and robust benchmark for Image Difference Captioning, covering ten difference categories across 1{,}075 compositional image pairs. To enable reliable evaluation, we introduce an \emph{LLM-as-a-Judge} protocol grounded in human-validated Difference Lists, measuring Recall$^{*}$, Precision$^{*}$, $F_{1}^{*}$, and Hallucination Rate. Extensive experiments on 16 MLLMs reveal a substantial performance gap between proprietary and open-source models, consistent improvements from reasoning-enabled variants, and persistent challenges in recognizing action, pose, and expression changes. Moreover, stable cross-judge rankings, strong agreement with human judgments, and high correlation with downstream image-editing performance demonstrate that DiffCap-Bench provides a reliable evaluation framework for IDC and serves as a practical proxy for MLLM utility in automated image-editing data engineering. Beyond leaderboard ranking, these results highlight that faithful difference understanding remains an open challenge for current MLLMs, especially under non-rigid and fine-grained visual changes. We hope DiffCap-Bench will foster more faithful evaluation of MLLMs and facilitate the development of stronger difference-aware data construction pipelines.

%% file: appendix/appendix.tex

\input{appendix/sections/overview}

\input{appendix/sections/DiffCap_BenchConstruct}

\input{appendix/sections/DimensionClassification}

\input{appendix/sections/DiffCap_Evaluation}

\input{appendix/sections/BenchmarkEffectiveness}


\begin{table*}[!htbp]
\centering
\begin{tcolorbox}[
    arc=4pt,
    boxrule=1pt,
    colback=gray!10,
    colframe=black,
    boxsep=0pt,
    left=4pt,
    right=4pt,
    width=\linewidth,
]
\label{tab:mllm_filter_prompt}

You are an expert visual analyst. Your task is to evaluate whether a pair of images should be retained for a visual difference benchmark. Base your judgment strictly on visible evidence; do not speculate beyond what can be observed.

\textbf{Evaluate the following criteria:}
\begin{enumerate}
    \item Difference Existence: determine whether there are real, observable differences between the two images.
    - Output ``Yes'' if there are clear, meaningful visual differences (e.g., object addition/removal, attribute change, spatial change).
    - Output ``No'' if the images are identical or only differ in negligible ways (e.g., compression noise, minor color fluctuation, tiny pixel-level variations).
    \item Visual Similarity: determine whether the two images are semantically related.
    - Output ``High'' if they depict the same scene or highly similar content with localized modifications.
    - Output ``Low'' if they depict different scenes or are completely unrelated (e.g., different objects, environments, or contexts).
    \item Image Quality: evaluate whether both images are of sufficient quality for reliable visual comparison.
    - Output ``Good'' if both images satisfy all of the following:
        (1) sufficient resolution (no extreme downsampling or tiny size),
        (2) clear and sharp content (no heavy blur or defocus),
        (3) objects are recognizable and not obscured,
        (4) no severe artifacts (e.g., compression blocks, corruption, overexposure).
    - Output ``Poor'' if either image violates any of the above conditions (e.g., blurry, low-resolution, heavily noisy, occluded, or visually ambiguous).
\end{enumerate}

\textbf{Guidelines:}
\begin{enumerate}
    \item Focus only on visible content; do not assume hidden or implied information.
    \item Ignore insignificant pixel-level differences when judging Difference Existence.
    \item Consider images similar only if they share the same scene or core semantic content.
    \item Treat completely unrelated image pairs as ``Low'' in Visual Similarity regardless of quality.
    \item Reject the pair if either image is difficult to interpret due to poor clarity or artifacts.
    \item Provide a concise explanation to justify your decision.
\end{enumerate}

\textbf{Output Format:}

Return your answer strictly in JSON format:

\{
``difference\_existence'': ``Yes/No'', \\
``visual\_similarity'': ``High/Low'', \\
``image\_quality'': ``Good/Poor'', \\
``reason'': ``brief explanation''
\}

\end{tcolorbox}
\caption{Prompt template for MLLM-based quality filtering of image pairs.}
\label{prompt_mllm_filter}
\end{table*}

\begin{table*}[!htbp]
\centering
\begin{tcolorbox}[
    arc=4pt,
    boxrule=1pt,
    colback=gray!10,
    colframe=black,
    boxsep=1pt,
    left=3pt,
    right=3pt,
    top=2pt,
    bottom=2pt,
    width=\linewidth,
]
\scriptsize
\setlength{\itemsep}{0.15em}
\setlength{\parsep}{0pt}
\setlength{\parskip}{0.2em}
You are an expert visual analyst. Your task is to classify a single described visual difference between two images into one or more of the following ten categories, allowing multi-label assignments when a difference involves multiple aspects. Base your judgment strictly on the provided description.

\textbf{Difference Categories:}
\begin{enumerate}
    \item \textbf{Object}: Additions, deletions, replacements, or changes in the number or type of objects and entities (e.g., a phone is removed; a person is added).
    \item \textbf{Attribute}: Changes in a visual property of an object or region, including color, size, shape, texture, material, clothing, or hair (e.g., the bottle becomes white; the cap is shorter).
    \item \textbf{Action \& Pose}: Changes in a subject's body motion or static posture (e.g., the person raises their arm; the subject shifts from sitting to standing).
    \item \textbf{Spatial}: Changes in the position, orientation, or relative arrangement of objects within the scene (e.g., the object moves left; two items swap positions).
    \item \textbf{Composition}: Changes in overall framing, subject scale, or cropping that alter the spatial layout of the scene (e.g., the subject is enlarged; the image is cropped more tightly).
    \item \textbf{Text \& Symbol}: Additions, deletions, or modifications to any text, logo, or symbolic element (e.g., the title text is changed; a logo is removed).
    \item \textbf{Background}: Changes to background content or the scene environment (e.g., the background becomes a solid color; the setting changes from indoors to outdoors).
    \item \textbf{Camera/View}: Changes in camera angle, shooting perspective, or lens parameters (e.g., the viewpoint shifts to overhead; the lens zooms in).
    \item \textbf{Image Property}: Global image-level changes not tied to any specific object, including resolution, aspect ratio, brightness, or saturation (e.g., overall saturation increases; the aspect ratio changes).
    \item \textbf{Expression}: Changes in a subject's facial expression or emotional state (e.g., the person smiles; the expression becomes serious).
\end{enumerate}
\textbf{Input:} A single difference description: ``\texttt{<difference description>}''

\textbf{Output Format:} Return your answer strictly in JSON format:
\{``types'': [``<category name>'', ...], ``reason'': ``<brief justification>''\}
\end{tcolorbox}
\caption{Prompt template for difference dimension classification.}
\label{prompt_diff_classify}
\end{table*}

\begin{table*}[!htbp]
\centering
\begin{tcolorbox}[
    arc=4pt,
    boxrule=1pt,
    colback=gray!10,
    colframe=black,
    boxsep=0pt,
    left=4pt,
    right=4pt,
    width=\linewidth,
]
\small
\label{tab:visual_change_prompt}

You are a visual change description model. Your task is to describe the observable changes required to transform an original image into a target image. Base your description only on visible evidence; do not invent objects, actions, or changes. You will also be provided with the resolution of each image. If the resolution has changed, you must mention it in your description. \\

\textbf{Include meaningful changes such as:}
\begin{enumerate}
    \item Objects: added, removed, repositioned, resized, recolored, or otherwise altered.
    \item Spatial and composition: layout, alignment, framing, cropping, or perspective changes.
    \item Camera and viewpoint: angle, zoom, or focal length changes.
    \item Image properties: resolution, sharpness, blur, noise, aspect ratio.
    \item Color, lighting, and style: tone, contrast, shadows, highlights, or overall style.
    \item Background and environment: visible changes in scene context or background.\\
\end{enumerate}

\textbf{Guidelines:}
\begin{itemize}
    \item Describe only changes, not what stays the same.
    \item Use clear, factual language in an editing-instruction style.
    \item Mention resolution changes if they occur.
    \item Do NOT refer to the images as ``Image A'' or ``Image B''.
    \item Enumerate multiple changes if necessary.
    \item Do not assume hidden content or speculate on intent.
    \item If a change is not clearly observable, do not mention it.\\
\end{itemize}

\textbf{Input:} \\

\textbf{Original Image Resolution} \\
\{\} \\

\textbf{Target Image Resolution} \\
\{\}

\end{tcolorbox}
\caption{Prompt template for evaluating image difference captioning in DiffCap-Bench.}
\label{prompt_visual_change}
\end{table*}

\begin{table*}[!htbp]
\centering
\begin{tcolorbox}[
    arc=4pt,
    boxrule=1pt,
    colback=gray!10,
    colframe=black,
    boxsep=0pt,
    left=4pt,
    right=4pt,
    width=\linewidth,
]
\label{tab:visual_change_evaluator}
You are a \textbf{strict, rational evaluator whose only source of truth is the visual evidence in the images}.
Your task is \textbf{NOT} to generate descriptions, but to evaluate how accurately the model-predicted change descriptions reflect the real visual changes from the original image to the target image. \\

\textbf{Task Background:} \\
The visual editing change description task is defined as follows:
Given an original image and a target image, a model must describe, in natural language, the visual editing operations required to transform the original into the target. You will receive two types of input:
\begin{enumerate}
    \item Model-predicted change descriptions (Prediction)
    \item Expert-annotated reference answers (Keypoint List)
\end{enumerate}

Your goal is to \textbf{systematically verify the prediction under the constraint of the reference answers and produce structured outputs for automatic evaluation}.\\
\\
\textbf{Input Description:} \\
\textbf{Model-Predicted Change Descriptions}:
May be completely correct; May mention real changes with incorrect details; May include additional valid changes; May miss some real changes; May contain hallucinations (non-existent or unchanged content)\\
\textbf{Keypoint List (Expert Annotation)}: (1) Key Change List: certain and important changes; one item may include multiple equivalent expressions separated by ``/'', all referring to the same underlying change. (2) Indistinguishable Items: changes annotators cannot fully confirm or are subjective; hitting these changes is \textbf{not} considered hallucination.

\textbf{Your Evaluation Task (Very Important):} \\
Evaluation consists of \textbf{Forward Checking} and \textbf{Backward Checking}. \\
\textbf{1. Forward Checking (Key-change-centered):} \\
Iterate through the Key Change List and evaluate each key change item. Assign one of the following hit statuses: (1) Hit and Correct: prediction mentions the corresponding change accurately (object, attribute, direction, quantity, spatial relation, etc.) (2) Hit but Incorrect: refers to the same change category but contains incorrect attributes (e.g., wrong color, spatial direction, object, quantity) (3) Not Hit: prediction does not mention the change or describes unrelated changes
\textit{Note: exact wording is not required; semantic equivalence is sufficient.} \\
\textbf{2. Backward Checking (Prediction-centered):} \\
Examine predicted descriptions that do \textbf{not} hit any key change item. Perform semantic decomposition:
(1) Treat each independent change as a minimal evaluation unit
(2) Split multiple changes in one description into separate units

\textbf{3. Extra Description Status Determination:} \\
For each extra change, assign: (1) Matches Indistinguishable Items: corresponds to an indistinguishable item (2) Extra Correct Description: truly exists from original to target but not recorded in keypoints
(3) Hallucination: describes non-existent or unchanged changes, or clearly contradicts visual facts

\textit{Important: ``Not in keypoint list'' does not imply hallucination. Resolution-related unchanged descriptions are also not hallucination unless indicated.} \\

\textbf{Output Requirements (Strict):} \\
Output \textbf{ONLY the final JSON result}. No reasoning or extra text. \\
The JSON structure must be:

\begin{verbatim}
{
  "Key_Change_List": [
    {"Description": "...", "Type": "...", "Hit_Status": "...", "Reason": "..."}
  ],
  "Extra_Description_List": [
    {"Description": "...", "Status": "...", "Reason": "..."}
  ],
  "Hallucination": {"Count": ..., "Reason": "..."}
}
\end{verbatim}

\textbf{Model-Predicted Difference Descriptions:  <prediction>}  \\\\
\textbf{Keypoint List: <difference list>} \\

\end{tcolorbox}
\caption{Prompt template used by the judge model to assess model-predicted difference descriptions.}
\label{prompt_visual_evaluator}
\end{table*}

\begin{figure*}[!t]
  \centering
\includegraphics[width=\textwidth,height=0.85\textheight,keepaspectratio]{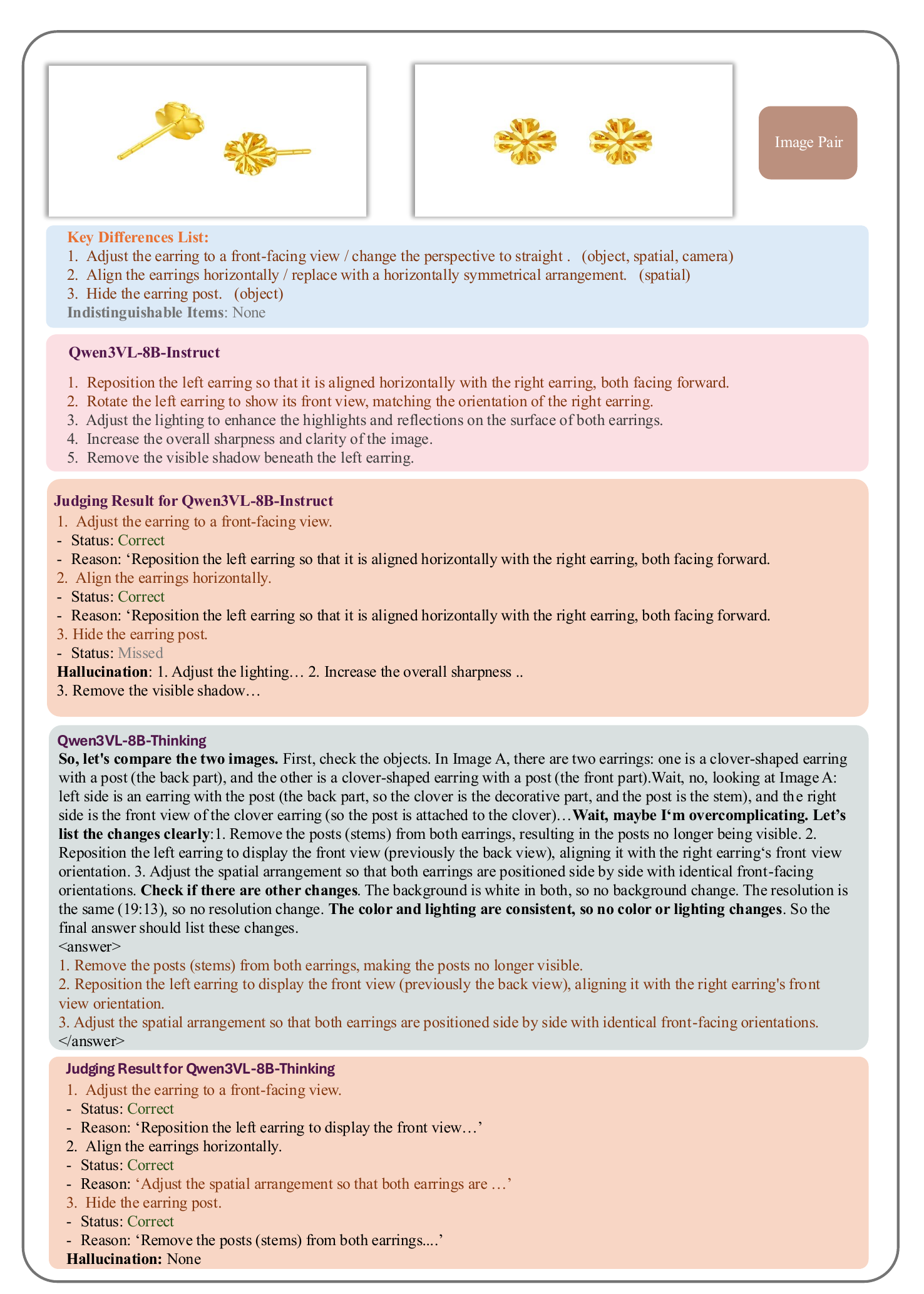}
\caption{Qualitative case study comparing Qwen3VL-8B-Instruct and Qwen3VL-8B-Thinking on the same sample, evaluated using the judge model on DiffCap-Bench. Qwen3VL-8B-Instruct outputs a list of differences directly, missing one true difference and producing three hallucinations. In contrast, Qwen3VL-8B-Thinking employs a think-before-output chain-of-thought (CoT) reasoning process, iteratively reflecting and self-checking to accurately identify all differences without hallucinations. This illustrates how CoT reasoning can improve the model's detection and description of visual differences.}
  \label{fig:case_study}
\end{figure*}

\begin{figure*}[t]
    \centering
    \includegraphics[width=\textwidth]{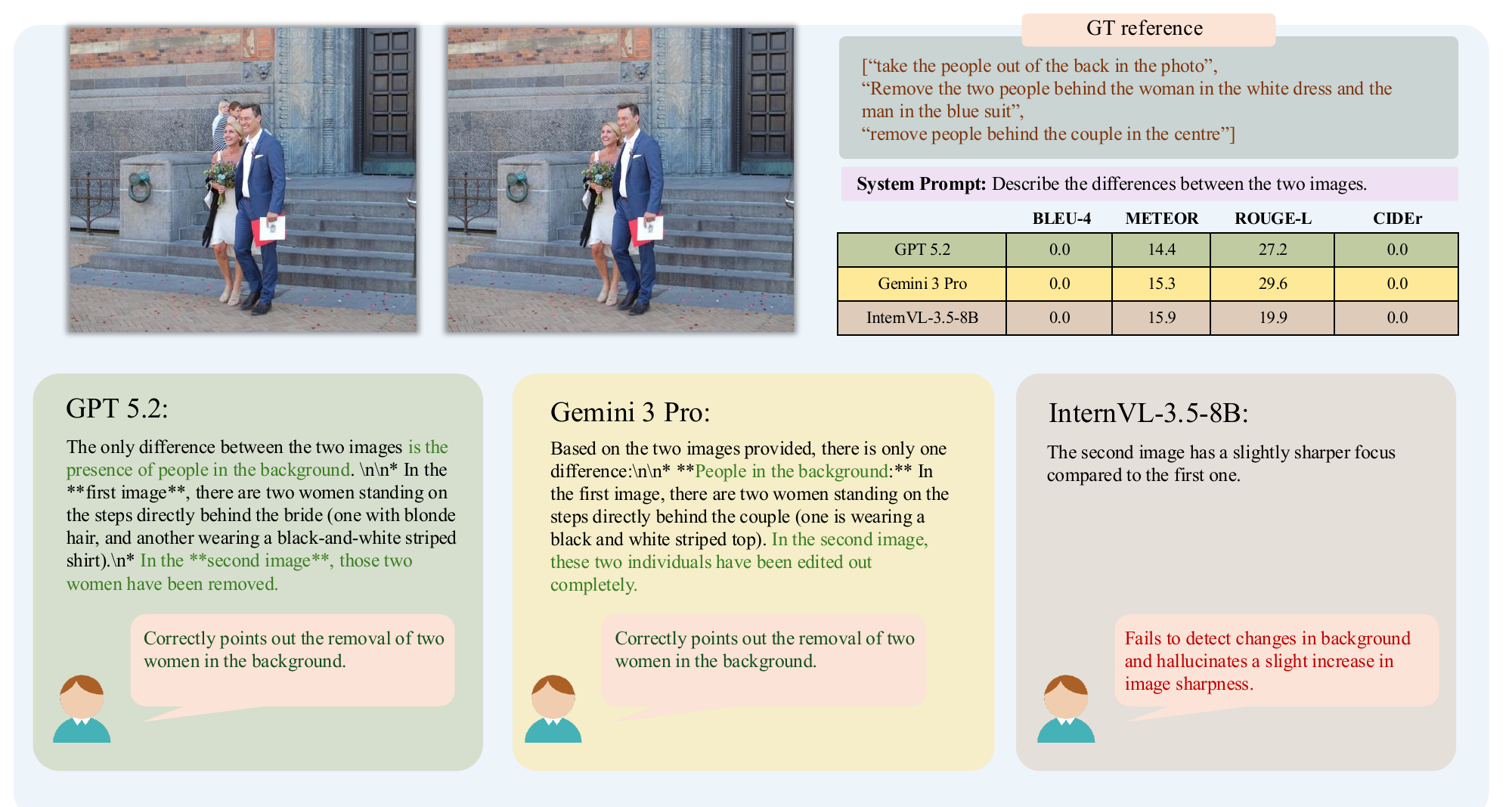}
\caption{Qualitative evaluation of MLLMs on the Image-Edit-Request benchmark using the same prompt template. GPT 5.2 and Gemini 3 Pro correctly identify the key difference—a person removed from the background—while InternVL3.5-8B fails to detect this change and introduces a hallucination by describing the image as ``slightly sharper''. Despite the stronger models producing more accurate and descriptive outputs, all models score poorly on traditional lexical-overlap metrics due to linguistic discrepancies with the reference captions, highlighting the limitations of conventional benchmarks for robust zero-shot assessment of MLLMs.}
    \label{fig:zeroshotIER}
\end{figure*}

%% file: appendix/sections/overview.tex
\section{Overview}
\label{sec:overview}
\begin{itemize}
    \item \textbf{Image Pair Quality Filtering Details} details the image pair quality filtering procedure, covering the prompt template for MLLM-based automatic filtering and the criteria used in human expert challengingness filtering.
    \item \textbf{Difference Dimension Classification} details the difference dimension classification procedure, including the classification prompt template.
    \item \textbf{Evaluation Details} presents the evaluation details on DiffCap-Bench, including the inference prompt template, inference parameters, judge prompt template, and a representative case study.
    \item \textbf{Benchmark Effectiveness} presents the evaluation details on traditional benchmarks along with qualitative examples, details the human alignment experiment (including the scoring rubrics for each evaluation criterion and the win rate calculation methodology), and the training details for the downstream image-editing correlation experiment.
\end{itemize}

%% file: appendix/sections/DiffCap_BenchConstruct.tex
\section{Image Pair Quality Filtering Details}
\label{imagepairqualityfiltering}

To ensure both data quality and benchmark challenge, we adopt a two-stage filtering procedure for all collected image pairs: automatic filtering using an MLLM to remove low-quality or invalid pairs, followed by human expert filtering to ensure sufficient challengingness.

\noindent\textbf{MLLM-Based Automatic Filtering. }
As shown in Table~\ref{prompt_mllm_filter}, we design a unified prompting protocol for Qwen3VL-Plus~\cite{qwen3vl} that evaluates each image pair across three criteria: \textit{difference existence}, \textit{visual similarity}, and \textit{image quality}. A pair is retained only if all three criteria are satisfied simultaneously (\textit{difference existence} = ``Yes'', \textit{visual similarity} = ``High'', \textit{image quality} = ``Good''); any pair failing on even one criterion is discarded. Pairs passing this automatic stage are then forwarded to human annotators for a second-round review.

\noindent\textbf{Human Expert Challengingness Filtering. }
Pairs that survive after filtering are further reviewed by human annotators to ensure sufficient challengingness. Annotators first enumerate all clearly observable differences. Let $N$ denote the total number of valid differences identified. If $N > 2$, the pair is directly retained. If $N \leq 2$, annotators additionally evaluate the relative visual area of each difference (i.e., the proportion of image pixels affected). Pairs in which any single difference occupies more than 20\% of the total image area are discarded, as such large and salient changes are perceptually easy to spot. This area-based criterion ensures that retained pairs contain only compact, localized differences that are inherently more challenging to detect.

%% file: appendix/sections/DimensionClassification.tex
\section{Difference Dimension Classification}
\label{dimension classification}
We employ Qwen3VL-Plus~\cite{qwen3vl} to categorize each annotated difference into predefined dimensions using the prompt template in Table~\ref{prompt_diff_classify}, allowing multi-label assignments since a single difference may involve multiple aspects. All dimension annotations are subsequently verified by human annotators to ensure label quality.

%% file: appendix/sections/DiffCap_Evaluation.tex
\section{Evaluation Details}
\label{diffcap_evaluation}
\noindent\textbf{Inference Details.} To evaluate the image difference captioning capability of MLLMs, we apply a unified prompt template (Table~\ref{prompt_visual_change}) across all models. Input images are constrained to a maximum resolution of $1024 \times 1024$ and a minimum resolution of $336 \times 336$. For all open-source models, we set the temperature to 0.1 and keep all other inference parameters at their default values. For all proprietary models, we use their default inference settings without any additional configuration.

\noindent\textbf{Judge Details.} For each sample in DiffCap-Bench, we employ Gemini 2.5 Pro~\cite{gemini} as the judge model, providing it with the image pair, the model's predicted descriptions, and the human-validated difference list as input (Table~\ref{prompt_visual_evaluator}). Based on the judging output, we calculate $\text{Recall}^*$, $\text{Precision}^*$, $\text{$F_{1}^{*}$}$, and the hallucination rate for each sample.

\noindent\textbf{Case Study.} We present a case study comparing Qwen3VL-8B-Instruct~\cite{qwen3vl} and Qwen3VL-8B-Thinking~\cite{qwen3vl} on the same sample, along with their respective judge outputs. As shown in Fig.~\ref{fig:case_study}, Qwen3VL-8B-Instruct~\cite{qwen3vl} directly produces a list of differences without any chain-of-thought reasoning, resulting in one missed difference and three hallucinations. In contrast, Qwen3VL-8B-Thinking~\cite{qwen3vl} follows a think-before-output CoT paradigm, carefully comparing the two images through iterative reflection and self-checking, ultimately identifying all differences correctly without introducing any hallucinations. This case illustrates that reasoning capability and the CoT paradigm can effectively enhance a model's ability to detect and describe visual differences, while also reducing hallucinations.

%% file: appendix/sections/BenchmarkEffectiveness.tex
\section{Benchmark Effectiveness}
\label{benchmarkeffectiveness}
\subsection{Evaluation on Traditional Benchmarks}
As shown in Fig.~\ref{fig:zeroshotIER}, we apply the same prompt template to evaluate MLLMs on two traditional image difference captioning benchmarks, Spot-the-Diff~\cite{spotdiff} and Image-Edit-Request~\cite{imgeditrequest}. Both GPT 5.2~\cite{gpt5} and Gemini 3 Pro~\cite{gemini} correctly identify that the key difference between the two images is the removal of a person in the background, whereas InternVL3.5-8B~\cite{internvl3d5} fails to detect this change and instead introduces a hallucination, describing the image as ``slightly sharper''. Nevertheless, despite the superior descriptive quality of the stronger models, all three models score poorly on lexical-overlap metrics due to linguistic discrepancies with the ground-truth reference sentences, exhibiting low separability across models. This demonstrates that traditional benchmark evaluation protocols are unable to provide robust zero-shot assessment of current MLLMs.

\subsection{Alignment with Human Judgment}
To validate the reliability of our automatic evaluation metrics, we conduct a human alignment study. For each image pair and its corresponding model-generated caption, human experts independently rate two aspects on a 1--5 scale: \textit{Holistic Quality} and \textit{Hallucination Severity}.

\noindent\textbf{Holistic Quality (1–5)} measures the model’s ability to accurately, comprehensively, and concisely describe all real differences between two images. A score of 5 (Excellent) indicates that all differences are observed and described correctly, with no errors. A score of 4 (Good) means that most differences are captured, allowing for at most one minor missed difference and at most one inaccurate description. A score of 3 (Fair) reflects that most differences are observed and described, but allows for up to two minor missed differences and up to two inaccurate descriptions. A score of 2 (Poor) indicates that only a few differences are captured or there are three or more significant errors in the descriptions. A score of 1 (Very Poor) means that no differences are captured or all descriptions are incorrect.

\noindent\textbf{Hallucination Severity (1--5)} assesses the extent to which the caption introduces content not present in the images, such as false objects, attributes, actions, or spatial arrangements. A score of 5 (None) indicates no hallucinated content. A score of 4 (Minimal) allows for a single hallucinated element that does not affect overall understanding. A score of 3 (Moderate) corresponds to exactly two hallucinated elements that may slightly mislead. A score of 2 (Severe) reflects three to four hallucinated elements that significantly misrepresent the differences. A score of 1 (Extreme) means five or more hallucinated elements, rendering the caption largely unrelated to the actual images.

\noindent\textbf{Evaluation Procedure.} We randomly sample 100 instances and collect outputs from six models: Gemini 3 Pro~\cite{gemini}, GPT 5.2~\cite{gpt5}, InternVL3.5-38B~\cite{internvl3d5}, Qwen3VL-32B-Thinking~\cite{qwen3vl}, Qwen3VL-8B-Thinking~\cite{qwen3vl}, and Qwen3VL-8B-Instruct~\cite{qwen3vl}. Each output is independently scored by six human annotators according to the criteria above.

\noindent\textbf{Win Rate Calculation.} For each sample $s$ and criterion $c$ (Holistic Quality or Hallucination Severity), the win rate of model $m$ is computed as:
\[
\text{WinRate}_{m,c} = \frac{\sum_{s=1}^{S} \sum_{\substack{m' = 1 \\ m' \neq m}}^{M} \mathbf{1}\big(f_{m,c}(s) > f_{m',c}(s)\big)}{S \cdot (M-1)}
\]
where $S$ is the total number of samples, $M$ is the total number of models, $f_{m,c}(s)$ is the score of model $m$ on sample $s$ under criterion $c$, and $\mathbf{1}(\cdot)$ is the indicator function that equals 1 if the condition holds and 0 otherwise. This formulation aggregates all pairwise comparisons across samples, providing a fine-grained measurement of relative model performance across the benchmark.

\subsection{Training Details}
\label{sec:trainingdetails}
We randomly sample 300K image pairs from Pico-Banana-400k~\cite{picobanana400k} and NanoBanana-150k~\cite{nanobanana150k}, two large-scale image editing datasets, as the training set, and 500 pairs as a held-out test set. Five MLLMs---InternVL3.5-8B~\cite{internvl3d5}, InternVL3.5-38B~\cite{internvl3d5}, Qwen3VL-8B-Instruct~\cite{qwen3vl}, Qwen3VL-8B-Thinking~\cite{qwen3vl}, and Qwen3VL-32B-Thinking~\cite{qwen3vl}---are used to relabel the training data, producing five distinct datasets with identical image pairs but different editing instructions. The base model \texttt{FLUX.1-Kontext-dev}~\cite{flux} is then fine-tuned separately on each relabeled dataset using supervised fine-tuning (SFT). To ensure a controlled comparison, we maintain a unified training setting: one epoch, a learning rate of $1\times 10^{-5}$, a total batch size of 32, freezing all other parameters and optimizing only the Diffusion Transformer (DiT) module. The resulting models are evaluated separately on the 500-pair test set following the ImgEdit~\cite{imgedit} evaluation protocol.